\documentclass[letterpaper]{article} % DO NOT CHANGE THIS
\usepackage[utf8]{inputenc} % Allow UTF-8 Unicode characters
\usepackage{aaai2026}  % DO NOT CHANGE THIS
\usepackage{times}  % DO NOT CHANGE THIS
\usepackage{helvet}  % DO NOT CHANGE THIS
\usepackage{courier}  % DO NOT CHANGE THIS
\usepackage[hyphens]{url}  % DO NOT CHANGE THIS
\usepackage{graphicx} % DO NOT CHANGE THIS
\usepackage{natbib}  % DO NOT CHANGE THIS AND DO NOT ADD ANY OPTIONS TO IT
\usepackage{caption} % DO NOT CHANGE THIS AND DO NOT ADD ANY OPTIONS TO IT
\newtheorem{theorem}{Definition}
\usepackage{algorithm}
\usepackage{algorithmic}

\usepackage{multirow}
\usepackage{multicol}
\usepackage{booktabs}
\usepackage{subfigure}
\usepackage{amssymb}
\usepackage{amsmath}
\usepackage[table]{xcolor}
\usepackage{xspace} 
\newcommand{\ours}{\texttt{MIP-Editor}\xspace}
\usepackage{newfloat}
\usepackage{listings}
\DeclareCaptionStyle{ruled}{labelfont=normalfont,labelsep=colon,strut=off} % DO NOT CHANGE THIS
\floatstyle{ruled}
\newfloat{listing}{tb}{lst}{}
\floatname{listing}{Listing}
\title{Cross-Modal Unlearning via Influential Neuron Path Editing in Multimodal Large Language Models}
\author{
    Kunhao Li\equalcontrib\textsuperscript{\rm 1},
    Wenhao Li\equalcontrib\textsuperscript{\rm 1},
    Di Wu\equalcontrib\textsuperscript{\rm 2},
    Lei Yang\textsuperscript{\rm 1}\thanks{Corresponding author. Email: \texttt{sely@scut.edu.cn}},
    Jun Bai\textsuperscript{\rm 3, 4},
    Ju Jia\textsuperscript{\rm 5},
    Jason Xue\textsuperscript{\rm 6}
}
\affiliations{
    \textsuperscript{\rm 1}School of Software Engineering, South China University of Technology, Guangzhou, China\\
    \textsuperscript{\rm 2}School of Computing, Engineering and Mathematical Science, La Trobe University, Melbourne, Australia\\
    \textsuperscript{\rm 3}School of Computer Science, McGill University, Montreal, Canada\\
    \textsuperscript{\rm 4}Mila-Quebec AI Institute, Montreal, Canada\\
    \textsuperscript{\rm 5}School of Cyber Science and Engineering, Southeast University, Nanjing, China\\
    \textsuperscript{\rm 6}CSIRO's Data61 and Responsible AI Research (RAIR) Centre, Adelaide University\\
    kunhomlihf@gmail.com, wenhaoli-lwh@outlook.com, d.wu@latrobe.edu.au, \\
    sely@scut.edu.cn, jun.bai@mcgill.ca, jiaju@seu.edu.cn, minhuixue@gmail.com\\
}

\usepackage{bibentry}
\begin{document}
\maketitle

\begin{abstract}
Multimodal Large Language Models (MLLMs) extend foundation models to real-world applications by integrating inputs such as text and vision. However, their broad knowledge capacity raises growing concerns about privacy leakage, toxicity mitigation, and intellectual property violations. Machine Unlearning (MU) offers a practical solution by selectively forgetting targeted knowledge while preserving overall model utility.
When applied to MLLMs, existing neuron-editing-based MU approaches face two fundamental challenges: (1) forgetting becomes inconsistent across modalities because existing point-wise attribution methods fail to capture the structured, layer-by-layer information flow that connects different modalities; and (2) general knowledge performance declines when sensitive neurons that also support important reasoning paths are pruned, as this disrupts the model’s ability to generalize.
To alleviate these limitations, we propose a multimodal influential neuron path editor (\ours) for MU. Our approach introduces modality-specific attribution scores to identify influential neuron paths responsible for encoding forget-set knowledge and applies influential-path-aware neuron-editing via representation misdirection. This strategy also enables effective and coordinated forgetting across modalities while preserving the model's general capabilities. Experimental results demonstrate that \ours achieves a superior unlearning performance on multimodal tasks, with a maximum forgetting rate of $87.75\%$ and up to $54.26\%$ improvement in general knowledge retention. On textual tasks, \ours achieves up to $80.65\%$ forgetting and preserves $77.9\%$ of general performance. Codes are available at \url{https://github.com/PreckLi/MIP-Editor}.
\end{abstract}

\section{Introduction}
\label{chap:intro}
The rapid advancement of multimodal large language models (MLLMs) has extended model capabilities to a wide range of applications through multimodal integration~\cite{2023llmsurvey,2024mllmsurvey}. However, their vast knowledge capacity raises serious concerns about privacy leakage~\cite{2024privacyconcern}, intellectual property violations~\cite{2024copyrightconcern}, regulatory compliance beyond privacy~\cite{chundawat2023zero}, toxicity mitigation~\cite{luckiadversarial}, and model refinement~\cite{jia2023model}. Machine Unlearning (MU)~\cite{2023llmunlearning} offers a promised solution to remove unwanted knowledge from MLLMs, supporting controllable and compliant model adaptation. However, current research on MU for MLLMs remains underexplored. 

\begin{figure}
  \centering
        \includegraphics[scale=0.59]{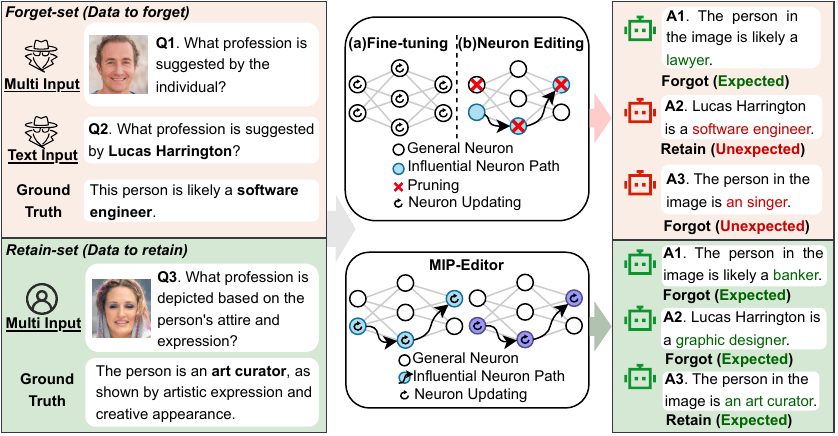}
  \caption{Comparison between existing MU methods and \ours. Prior methods suffer from: (1) insufficient forgetting in the text modality, as point-wise attribution fails to capture structured cross-layer information flow; and (2) disruption of influential reasoning paths due to pruning.}
  \label{fig:motivation}
\end{figure}

Methods such as~\cite{2022grad_ascend, 2022grad_diff, 2024npo} primarily extend fine-tuning-based unlearning strategies originally designed for LLMs. However, these methods ignore the unique discrepancies between modalities in MLLMs, and struggle to forget modality-specific knowledge effectively, especially under textual inputs, as illustrated in Fig.~\ref{fig:motivation} (a). An alternative line of work that explores neuron-level editing has emerged as a promising direction, based on the observation that model knowledge is stored in distributed patterns within learnable parameters~\cite{2023neuron_editing}. Recent approaches such as DEPN~\cite{2023depn} and MANU~\cite{2025manu} attempt modality-specific forgetting via single-neuron pruning or activation-based scoring. However, the point-wise estimation fails to capture the structured information flow across layers in multimodal architectures. As a result, forgetting remains uncoordinated across modalities. This limitation is statistically demonstrated in Table~\ref{tab:result_5}. Moreover, as illustrated in Fig.~\ref{fig:motivation} (b), pruning neurons solely based on their individual importance to the forget-set may inadvertently disrupt reasoning pathways critical to the retain-set, leading to severe degradation of general knowledge.

% An alternative line of work that explores neuron-level editing has emerged as a promising direction, based on the observation that model knowledge is stored in distributed patterns within learnable parameters~\cite{2023neuron_editing}. Recent approaches such as DEPN~\cite{2023depn} and MANU~\cite{2025manu} attempt modality-specific forgetting via single-neuron pruning or activation-based scoring. However, the point-wise estimation fails to capture the structured information flow across layers in multimodal architectures. As a result, forgetting remains uncoordinated across modalities. This limitation is statistically demonstrated in Table~\ref{tab:result_5}. Moreover, as illustrated in Fig.~\ref{fig:motivation}, pruning neurons solely based on their individual importance to the forget-set may inadvertently disrupt reasoning pathways critical to the retain-set, leading to severe degradation of general knowledge.
% Furthermore, such localized edits often result in fragmented forgetting that fails to comprehensively erase semantic concepts.
Recent studies~\cite{2025influential_path} have confirmed that information in large models is transmitted through structured, layer-wise neuron pathways. These influential paths offer a more coherent and semantically grounded basis for unlearning compared to isolated neurons. In MLLMs, both textual and multimodal (e.g., image–text) inputs rely on such structured reasoning flows. This motivates a shift from point-based deletion to path-aware interventions that better align with the model’s internal knowledge organization. To this end, we propose a \textbf{M}ultimodal \textbf{I}nfluential neuron \textbf{P}ath \textbf{Editor} (\ours) tailored for MU in MLLMs. Our approach locates modality-specific influential neuron paths in the FFN layers of both the textual and visual branches by computing inter-layer gradient-integrated and Fisher-integrated attribution scores. In particular, we introduce an influential-path-based neuron editing method using Representation Misdirection Unlearning (RMisU) that adaptively steers the representations of forget-set inputs away from their original semantics, reducing the impact on general knowledge.
To sum up, our contributions are as follows:
\begin{itemize}
    \item We propose a dual-branch (visual and textual) influential neuron path localization framework. This approach leverages inter-layer gradient-integrated and Fisher-integrated attribution scores to capture modality-specific information flow, enabling precise localization of neurons responsible for specific knowledge in each modality.
    
    \item We analyze the limitations of direct pruning strategies, where overlapping neurons between forget and retain sets cause a collapse of general reasoning paths. To mitigate this, we propose a targeted RMisU-based neuron editing strategy that operates only on the influential neuron paths, decoupling specific and general knowledge.
    
    \item Experiments demonstrate that \ours achieves modality-consistent forgetting with strong retain-set performance, reaching up to $87.75\%$ forgetting and $54.26\%$ retention improvement on multimodal tasks, and $80.65\%$ forgetting with $77.9\%$ retention on textual tasks.
\end{itemize}

\section{Problem Definition}
In this work, we focus on MU for MLLMs, aiming to remove targeted forgetting knowledge while minimizing degradation of general capabilities. Let $M_{\theta}$ denote the original MLLM with parameters $\theta$, trained on a dataset $D = \{(I_i, T_i)\}_{i=1}^{\mathbf{N}}$ of $\mathbf{N}$ image–text pairs, where $I_i$ is an image and $T_i = \{s^i_1, \dots, s^i_{t_i}\}$ is its corresponding tokenized text. Each pair includes a question–answer prompt for visual understanding.
We divide $D$ into a forget-set $D^f = \{(I^f_j, T^f_j)\}_{j=1}^{\mathbf{N}_f}$, containing specific concepts to be forgotten, and a retain-set $D^r = \{(I^r_k, T^r_k)\}_{k=1}^{\mathbf{N}_r}$, used to preserve general knowledge.

Following~\cite{2025mllmubench, 2025manu}, we define MU in MLLMs as: \textit{The process of removing both visual and textual forgetting data from a model while preserving its predictive performance on unrelated inputs.}
To achieve this, we minimize the negative log-likelihood of next-token prediction and obtain the unlearned model $M_{\hat{\theta}}$ via the objective:
\begin{align}
    \nonumber
        &\mathop{\arg \min}_{\theta^*}\Biggl\{\underbrace{-\mathbb{E}_{(I,T)\in D^{f}}\Bigl[-\sum_{\mathbf{n}=1}^{\mathbf{N}_f}\log p_{M_{\hat{\theta}}}(w_{\mathbf{n}}|(I,T),\,w_{<\mathbf{n}})\Bigr]}_{\text{Forget specific visual \& textual patterns}}\\
        &+\underbrace{\mathbb{E}_{(I,T)\in D^{r}}\Bigl[-\sum_{\mathbf{n}=1}^{\mathbf{N}_r}\log p_{M_{\hat{\theta}}}(w_{\mathbf{n}}| (I,T),\,w_{<\mathbf{n}})\Bigr]}_{\text{Retain general knowledge}}
        \Biggr\}
\end{align}

\section{Method}
% We propose \ours, a two-stage unlearning framework for MLLMs that removes privacy-sensitive knowledge while preserving general capabilities, as illustrated in Fig.~\ref{fig:framework}.
% We propose \ours, a two-stage unlearning framework for MLLMs that targets privacy-sensitive knowledge while preserving general capabilities. As Fig.~\ref{fig:framework} shows, the first stage locates modality-specific influential neuron paths. The second stage applies path-specific fine-tuning, enabling effective knowledge removal with minimal impact on the retain-set. Algorithm~\ref{alg:overall_framework} also shows the overall process of \ours.
\begin{figure*}
  \centering
        \includegraphics[scale=0.8]{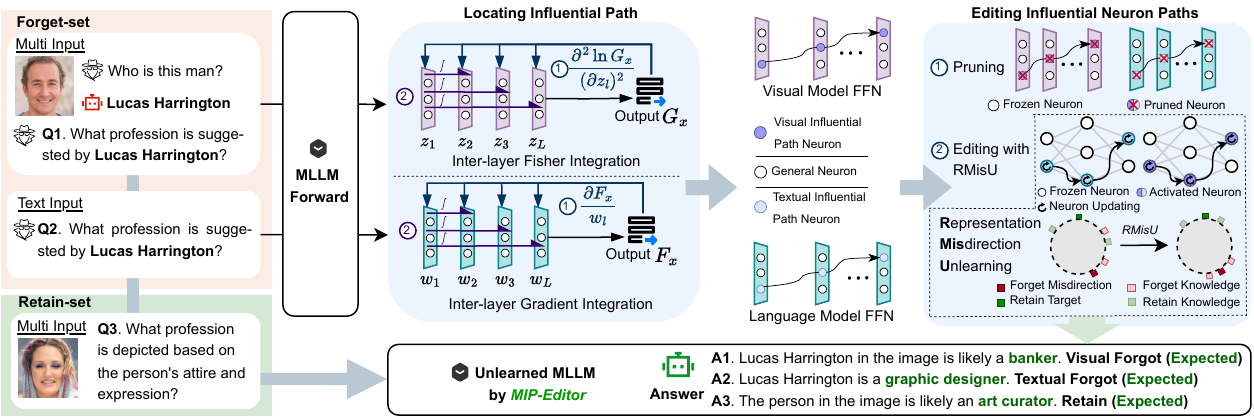}
  \caption{Overview of \ours. (1) Influential neuron paths are located using inter-layer gradient (text) and Fisher (vision) integration. (2) Neurons inside the selected paths are pruned, and (3) path-specific editing is performed via representation misdirection to achieve modality-consistent forgetting while preserving general knowledge.}
  \label{fig:framework}
\end{figure*}
\subsection{Locating Influential Neuron Path}
In \ours (Fig.~\ref{fig:framework}), the textual and visual influential paths are inherently related, as each input pair $(I, T)$ carries semantically aligned content. Prior works~\cite{2021clip, 2024finding, 2025identifying} show that vision and text features are mapped into a shared embedding space and jointly processed via cross-attention, ensuring correspondence between activations across modalities. 
% This alignment enables the layer-wise path locating in \ours to identify modality-specific neurons that are semantically consistent.
% In MLLMs, learnable parameters are divided between the visual and language models. To ensure consistent forgetting across modalities, we propose two modality-specific neuron importance metrics: an inter-layer gradient integration score for textual neurons and an inter-layer Fisher integration score for visual neurons, enabling independent localization of influential neuron paths in each modality.

\subsubsection{Inter-layer Gradient Integration}
To locate influential neurons in the textual modality, we propose an inter-layer gradient integration method inspired by information flow~\cite{2021infomation_flow} and joint attribution~\cite{2025influential_path}. Given the $L$-layer architecture of a language model, where each FFN layer is regarded as a key repository of factual knowledge, we aim to quantify the contribution of selected neurons across the first $N$ layers.

Let $\langle T, Y \rangle$ denote a labeled text pair, where $T \in \mathbb{R}^d$ is the input text and $Y$ is the expected output. The model’s output over the first $N$ layers is represented as:
\begin{equation}
    F_T(\mathbf{w}) = p(Y \mid T, w_{i_1}^1, \cdots, w_{i_N}^N),
\end{equation}
where $\mathbf{w} = (w_{i_1}^1, \cdots, w_{i_N}^N)$ are the activations of selected neurons in the textual FFN layers, and $\tilde{w}_{i_n}^n$ denotes the original activation of neuron $w_{i_n}^n$ of the $n$-th layer.

To estimate the joint contribution of these neurons, we scale the activation values $\{\alpha_{i_1}^1, \cdots, \alpha_{i_N}^N\}$ from 0 to their original activations $\{\tilde{w}_{i_1}^1, \cdots, \tilde{w}_{i_N}^N\}$. The inter-layer gradient-integrated attribution score is defined as:
\begin{equation}
    \text{IGI}(\mathbf{w}) = \sum_{n=1}^N \tilde{w}_{i_n}^n \int_0^{\tilde{w}_{i_n}^n} \sum_{l=1}^N \frac{\partial F_T(\alpha_{i_1}^1, \cdots, \alpha_{i_n}^n)}{\partial w_{i_l}^l} \, d\alpha_{i_n}^n,
\end{equation}
which measures how the neurons along the path contribute to the model's output by integrating gradients across layers.

To approximate the integral, we employ Riemann approximation~\cite{2022riemann} by interpolating $m$ frames into the activation values. The discrete form of the IGI becomes:
\begin{equation}
    \label{eq:igi}
    \text{IGI}(\mathbf{w}) = \sum_{j=1}^N \tilde{w}_{i_j}^n \sum_{k=1}^m \sum_{l=1}^N \frac{\partial F_T\left(\frac{k}{m} \alpha_{i_1}^1, \cdots, \frac{k}{m} \alpha_{i_N}^N\right)}{\partial w_{i_l}^l}.
\end{equation}
\subsubsection{Inter-layer Fisher Integration}
To locate influential neurons in the visual modality, we adopt an inter-layer Fisher integration method similar to the gradient-based approach used for text. Due to the high dimensionality, spatial correlation, and parameter redundancy in vision encoders, the Fisher Information Matrix (FIM) offers a more suitable signal for estimating neuron importance.

Let $\langle (I, T), Y \rangle$ denote a multimodal input with visual input $I \in \mathbb{R}^{d^I}$, text input $T \in \mathbb{R}^{d^T}$, and target output $Y$. The log-likelihood output over the first $N$ visual FFN layers is:
\begin{equation}
    \mathbf{G}(\mathbf{z}) = \log p(Y \mid I, T, z_{i_1}^1, \cdots, z_{i_N}^N),
\end{equation}
where $\mathbf{z} = (z_{i_1}^1, \cdots, z_{i_N}^N)$ are the activations of selected visual neurons, and $\tilde{z}_{i_n}^n$ their original values.

Similar to the textual integration (Eq.~\ref{eq:igi}), we interpolate neuron activations from 0 to their original values using $m$ steps. To approximate the diagonal of the FIM, we adopt the squared-gradient formulation. The inter-layer Fisher-integrated score is defined as:
\begin{equation}
    \label{eq:ifi}
    \text{IFI}(\mathbf{z}) = \sum_{n=1}^N \tilde{z}_{i_n}^n \sum_{k=1}^m \sum_{l=1}^N \left( \frac{\partial \mathbf{G}\left(\frac{k}{m} \beta_{i_1}^1, \cdots, \frac{k}{m} \beta_{i_N}^N \right)}{\partial z_{i_l}^l} \right)^2,
\end{equation}
where $\{\beta_{i_n}^n\}$ are the interpolated activations. This formulation enables efficient estimation of visual neuron importance by integrating second-order signals across layers.
\subsubsection{Locating Paths}

Following~\cite{2025influential_path}, we define the influential neuron paths in the FFN layers of MLLMs.

\begin{theorem}[Influential Paths]
Let $F: \mathbb{R}^d \to \mathbb{R}$ be a multimodal model consisting of $L$ FFN layers for a given modality, and let $x$ denote either a text input $T$ or an image–text pair $(I,T)$. A neuron path is defined as
\begin{equation}
    \mathcal{P}_x = \{w^1, w^2, \dots, w^L\},\qquad w^l \in \mathbf{W},
\end{equation}
where $w_l$ denotes the set of selected neurons in the $l$-th layer. $\mathbf{W}$ represents the general FFN layer parameters

We define a modality-specific scoring function as follows:
\begin{equation}
    S(\mathcal{P}_x) = 
    \begin{cases}
    \mathrm{IGI}(\mathcal{P}_x), & \text{if } x = T, \\
    \mathrm{IFI}(\mathcal{P}_x), & \text{if } x = (I,T),
    \end{cases}
\end{equation}
The influential path $\mathcal{P}_x^*$ is then defined as the one that maximizes the corresponding score:
\begin{equation}
    \mathcal{P}_x^* = \arg\max_{\mathcal{P}_x} S(\mathcal{P}_x).
\end{equation}
\end{theorem}

To locate influential paths efficiently, we apply a greedy layer-wise search (Algorithm~\ref{alg:igi_ifi_path}) that selects the most influential neuron per layer. Given input $(I, T)$ and a pretrained model $F$ with $L^t$ textual and $L^v$ visual FFN layers, the algorithm outputs two ordered paths: $\mathcal{P}^t$ and $\mathcal{P}^v$ for the textual and visual modalities, respectively.

\begin{algorithm}[t]
\caption{Inter-layer Integrated Influential Path Locating}
\label{alg:igi_ifi_path}
\begin{algorithmic}[1]
\REQUIRE MLLM $M_{\theta}$ with $L^t$ textual layers and $L^v$ visual layers, input pair $(I, T)$
\ENSURE Visual path $\mathcal{P}^v$, Textual path $\mathcal{P}^t$

\STATE $\mathcal{P}^v \leftarrow \emptyset$, $\mathcal{P}^t \leftarrow \emptyset$

% \STATE \textbf{// Textual Influential Path Locating}
\FOR{$k = 1$ to $L^t$} 
    \STATE Let $\mathcal{W}^t$ be the set of neurons in textual layer $k$
    \STATE $\text{Score} \leftarrow -\infty$, $\text{bestNeuron} \leftarrow \text{None}$
    \FORALL{$w \in \mathcal{W}^t$}
        \STATE $s \leftarrow \mathrm{IGI}(\mathcal{P}^t \cup \{w\}, T)$
        \IF{$s > \text{Score}$}
            \STATE $\text{Score} \leftarrow s$, $\text{bestNeuron} \leftarrow w$
        \ENDIF
    \ENDFOR
    \STATE $\mathcal{P}^t \leftarrow \mathcal{P}^t \cup \{\text{bestNeuron}\}$
\ENDFOR

% \STATE \textbf{// Visual Influential Path Locating}
\FOR{$l = 1$ to $L^v$}
    \STATE Let $\mathcal{W}^v$ be the set of neurons in visual layer $l$
    \STATE $\text{Score} \leftarrow -\infty$, $\text{bestNeuron} \leftarrow \text{None}$
    \FORALL{$z \in \mathcal{W}^v$}
        \STATE $s \leftarrow \mathrm{IFI}(\mathcal{P}^v \cup \{z\}, I, T)$
        \IF{$s > \text{Score}$}
            \STATE $\text{Score} \leftarrow s$, $\text{bestNeuron} \leftarrow z$
        \ENDIF
    \ENDFOR
    \STATE $\mathcal{P}^v \leftarrow \mathcal{P}^v \cup \{\text{bestNeuron}\}$
\ENDFOR

\RETURN $\mathcal{P}^v,\ \mathcal{P}^t$
\end{algorithmic}
\end{algorithm}

\subsection{Editing Influential Neuron Paths}
Having located the influential neuron paths, we interrupt the encoded forgetting information by pruning these neurons and then apply RMisU to steer their activations away from the forget-set semantics while reinforcing retain-set representations. Unlike full-model retraining, this targeted prune‑and‑finetune strategy updates only a small subset of neurons, achieving effective forgetting with minimal impact on general knowledge.
% It also reduces memory overhead and speeds up convergence.

\subsubsection{Pruning}
Specifically, we perform targeted pruning by zeroing the activations of neurons identified as specific-relevant along the influential paths. For a text-only input $ T $ and a multimodal input $ (I, T) $, let the corresponding influential paths be 
$\mathcal{P}_T = \{\tilde{w}_1, \tilde{w}_2, \dots, \tilde{w}_{L^t}\}$ and $\mathcal{P}_{(I,T)} = \{\tilde{z}_1, \tilde{z}_2, \dots, \tilde{z}_{L^v}\}$, the pruning can be formally expressed as:
\begin{equation}
    \tilde{w}_l \leftarrow \mathbf{0},  \forall l \in \{1, \dots, L^t\}; \quad
\tilde{z}_l \leftarrow \mathbf{0}, \forall l \in \{1, \dots, L^v\},
\end{equation}
where $\mathbf{0}$ represents an all-zero vector with the same dimension, $\tilde{w}_l$ and $\tilde{z}_l$ denote the activation values set of the selected neurons in the $l$-th layer. This operation ensures that the flow of information associated with forgetting concepts is blocked, thereby achieving targeted forgetting.

\subsubsection{Editing with RMisU}
Pruning risks losing general knowledge in overlapping neurons of the retain-set. To recover it adaptively with minimal forgetting, we fine-tune only the pruned neurons using the retain-set, enabling adaptive recovery with less reintroduction of forgotten content.
Specifically, all other parameters in the MLLM are frozen, and only neurons along influential paths are updated. Let $M_{\theta^*}$ denote the pruned model. To preserve general knowledge, we minimize a cross-entropy loss over the retain-set $ D^r$:
\begin{equation}
    \mathcal{L}_{\text{retain}} = \mathbb{E}_{(x^r, y^r) \in D^r} \left[ - \sum_{i=1}^{|y^r|} \log P_{M_{\theta^*}}(y^r_i \mid x^r, y^r_{<i}) \right],
\end{equation}
where $ x^r \in \{T^r, (I^r, T^r)\} $ denotes either a text or image–text input, and $ y^r $ is the corresponding output sequence.

To forget specific knowledge in the forget-set $ D^f $, prior methods use gradient ascent, KL divergence, or contrastive objectives, but often at the cost of linguistic and utility degradation~\cite{2024simnpo}.
To avoid this, we adopt adaptive Representation Misdirection Unlearning (RMisU)~\cite{2025adaptivermu}, which steers forget-set representations away from their original semantics via localized directional perturbation at a specific layer $l$. This targeted editing removes specific knowledge while preserving general linguistic ability~\cite{2024rmu}.

For each forget-set input $x^f \in \{T^f, (I^f, T^f)\}$, we sample a random unit vector from the unit sphere:
\begin{equation}
    \mathbf{u} \sim \mathrm{Uniform}\left(\mathbb{S}^{d-1}\right),
\end{equation}
and define a layer-specific target representation as
\begin{equation}
    \label{eq:target_representation}
    \mathbf{v}^f = \lambda \cdot \left\|\mathbf{h}^{(l)}_{M_\theta}(x^f)\right\|_2 \cdot \mathbf{u},
\end{equation}
where $\mathbf{h}^{(l)}_{M_\theta}(x^f)$ denotes the frozen model's hidden representation at layer $l$, and $\lambda$ is a scaling coefficient modulating the influence of the perturbation.
\paragraph{Forgetting RMisU loss.}
This term forces the representation of forget-set samples to align with the randomized vector $\mathbf{v}_F$, effectively erasing forgetting knowledge:
\begin{equation}
    \label{eq:forget_rmisu}
    \mathcal{L}^f_{\mathrm{RMisU}} = \mathbb{E}_{x^f \in D^f}
    \left\|
        \mathbf{h}_{M_{\theta*}}^{(l)}(x^f)
        - \mathbf{v}^f
    \right\|_2^2,
\end{equation}
where $\mathbf{h}_{M_{\theta*}}^{(l)}(x^f)$ denotes the intermediate representation at layer $l$ for a forget-set sample $x^f$ in $M_{\theta^*}$ at current epoch.
\paragraph{Retaining RMisU loss.}
We minimize deviation of retain-set representations from the frozen model for generalization:
\begin{equation}
    \label{eq:retain_RMisU}
    \mathcal{L}^r_{\mathrm{RMisU}} = \mathbb{E}_{x^r \in D_{\mathrm{r}}}
    \left\|
        \mathbf{h}_{M_{\theta*}}^{(l)}(x^r)
        - \mathbf{h}_{M_\theta}^{(l)}(x^r)
    \right\|_2^2.
\end{equation}

\paragraph{Full objective.}
The overall adaptive RMisU loss is:
\begin{equation}
    \mathcal{L}_{\mathrm{RMisU}} = \mathcal{L}^f_{\mathrm{RMisU}} + \gamma \cdot \mathcal{L}^r_{\mathrm{RMisU}},
\end{equation}
where $\gamma > 0$ balances forgetting and retention.

\section{Experiments}
\begin{table*}
\centering
\setlength{\tabcolsep}{3.2pt}
\small
\caption{Overall performances of baseline methods and \ours on machine unlearning tasks with 5\% forget ratio. F: Forget-set; R:Retain-set; VQA:Vision Question Answer; QA: Question Answer; VGEN: Vision Generation; GEN: Generation.}
\begin{tabular}{c|cccccc|cccccc}
\toprule
Method    & \multicolumn{6}{c|}{MLLMU-Bench}                 & \multicolumn{6}{c}{CLEAR}                       \\
\midrule
Task      & FVQA & RVQA & FVGEN & RVGEN & FQA & RQA & FVQA & RVQA & FVGEN & RVGEN & FGEN & RGEN \\
Metric&Acc($\downarrow$)&Acc($\uparrow$)&Rouge($\downarrow$)&Rouge($\uparrow$)&Acc($\downarrow$)&Acc($\uparrow$)&Acc($\downarrow$)&Acc($\uparrow$)&Rouge($\downarrow$)&Rouge($\uparrow$)&Rouge($\downarrow$)&Rouge($\uparrow$)\\
\midrule
\multicolumn{13}{c}{Qwen2.5-VL-3B-Instruct}                                                                             \\
\midrule
Vanilla  & 39.20\% & 37.72\% & 0.4527 & 0.4347 & 49.60\% & 47.20\% & 72.34\% & 73.42\% & 0.3196 & 0.2997 & 0.3776 & 0.3900 \\
GA\_Diff & 32.00\% & 32.80\% & 0.4450 & 0.4756 & 46.40\% & 43.20\% & 27.66\% & 23.04\% & 0.2946 & 0.2751 & 0.3740 & 0.3896 \\
KL\_Min  & 33.60\% & 27.59\% & 0.2139 & 0.1940 & 41.60\% & 42.57\% & 12.77\% & 9.11\%  & 0.2532 & 0.2400 & 0.3270 & 0.3287 \\
NPO      & 37.60\% & 36.20\% & 0.4507 & 0.4307 & 42.40\% & 44.80\% & 7.45\%  & 9.37\%  & 0.0803 & 0.0605 & 0.0805 & 0.0639 \\
MANU     & 36.00\% & 34.47\% & 0.4406 & 0.4367 & 30.80\% & 34.65\% & 78.72\% & 77.97\% & 0.3220 & 0.2987 & 0.3809 & 0.3903 \\
% LLMEraser &       &        &   &     &       &     &     &    \\
\rowcolor{gray!30}\ours & 4.80\% & 58.19\% & 0.0997 & 0.4195 & 9.60\% & 36.80\% & 3.19\% & 24.05\% & 0.0707 & 0.2684 & 0.0926 & 0.3631 \\
\midrule
\multicolumn{13}{c}{Llava-1.5-7B}                                                                              \\
\midrule
Vanilla   & 56.80\% & 51.56\% & 0.5580 & 0.4946 & 50.40\% & 52.59\% & 44.68\% & 43.54\% & 0.3060 & 0.2937 & 0.3462 & 0.3546 \\
GA\_Diff  & 54.40\% & 52.78\% & 0.5719 & 0.5071 & 42.40\% & 49.83\% & 14.36\% & 15.19\% & 0.3057 & 0.2931 & 0.3565 & 0.3620 \\
KL\_Min   & 32.80\% & 38.27\% & 0.3594 & 0.3390 & 43.20\% & 43.29\% & 43.62\% & 42.28\% & 0.2200 & 0.2068 & 0.1380 & 0.1671 \\
NPO       & 48.00\% & 47.26\% & 0.5388 & 0.4907 & 46.40\% & 51.52\% & 10.64\% & 15.95\% & 0.2091 & 0.1815 & 0.0150 & 0.0136 \\
MANU      & 56.00\% & 52.11\% & 0.5486 & 0.4960 & 48.80\% & 52.19\% & 43.62\% & 42.28\% & 0.3070 & 0.2920 & 0.3452 & 0.3556 \\
% LLMEraser &     &     &    &     &    &      &     &    \\
\rowcolor{gray!30}\ours & 38.40\% & 47.22\% & 0.3418 & 0.3552 & 36.80\% & 47.34\% & 6.38\% & 52.66\% & 0.9690 & 0.2258 & 0.1441 & 0.2268 \\
\bottomrule
\end{tabular}
\label{tab:result_5}
\end{table*}
In this section, we answer the following key questions concerning the performance of \ours with experiments. \textbf{Q1: Can \ours effectively eliminate multimodal information from the target MLLMs? Q2: Can \ours achieve coordinated forgetting across visual and textual modalities? Q3: Can \ours strike a balance between forgetting information and preserving general knowledge? Q4: Does \ours retain more informative content through influential neuron paths compared to point-wise probing methods?}

\subsection{Experimental Setup and Baselines}
To evaluate the effectiveness of \ours, we conduct experiments on two representative MLLMs of different scales: Qwen2.5-VL-3B-Instruct~\cite{2024qwen2vl} and LLaVA1.5-7B~\cite{2023llava}, using two dedicated multimodal unlearning benchmarks: MLLMU-Bench~\cite{2025mllmubench} and CLEAR~\cite{2024clear}. These datasets provide structured forget and retain splits across diverse multimodal tasks, including visual question answering (VQA) and text-based QA, covering both generation and classification settings. We compare \ours with four strong baselines: GA\_Diff~\cite{2022grad_diff}, KL\_Min~\cite{2020kldivergence}, NPO~\cite{2024npo}, and MANU~\cite{2025manu}. Vanilla denotes the original model without unlearning. For fair comparison, all methods are trained using the same configurations. MLLMU-Bench uses 5\%, 10\%, and 15\% of its samples as forget-sets, while CLEAR uses 1\%, 5\%, and 10\%.

\subsection{Main Results}
To answer Q1 and Q2, we evaluate the unlearning performance of various methods on multimodal and textual tasks using Qwen2.5-VL-3B-Instruct and LLaVA1.5-7B under a 5\% forget ratio on MLLMU-Bench and CLEAR (Table~\ref{tab:result_5}). On multimodal tasks, \ours significantly reduces forgetting knowledge retention. For instance, it lowering FVQA accuracy from $39.20\%$ (Vanilla) to $4.80\%$ and improving RVQA from $37.72\%$ to $58.19\%$ on MLLMU based on Qwen2.5-VL. This corresponds to an $87.75\%$ forgetting rate and a $54.26\%$ improvement in general knowledge retention, outperforming GA\_Diff, KL\_Min, and NPO. Similar trends are observed for LLaVA1.5.
On textual tasks, \ours reduces FQA accuracy from $49.60\%$ to $9.60\%$, achieving an $80.65\%$ forgetting rate while retaining $77.9\%$ of the original performance. Compared with MANU, NPO, and KL\_Min, \ours more effectively suppresses residual forgetting knowledge while preserving competitive accuracy on the retain-set (e.g., $58.19\%$ RVQA and $47.34\%$ RQA). These results confirm the strength of \ours in achieving coordinated forgetting across modalities with minimal impact on general capability.

\subsection{Unlearning v.s. Model Utility}
\begin{figure}
  \centering
        \includegraphics[scale=0.32]{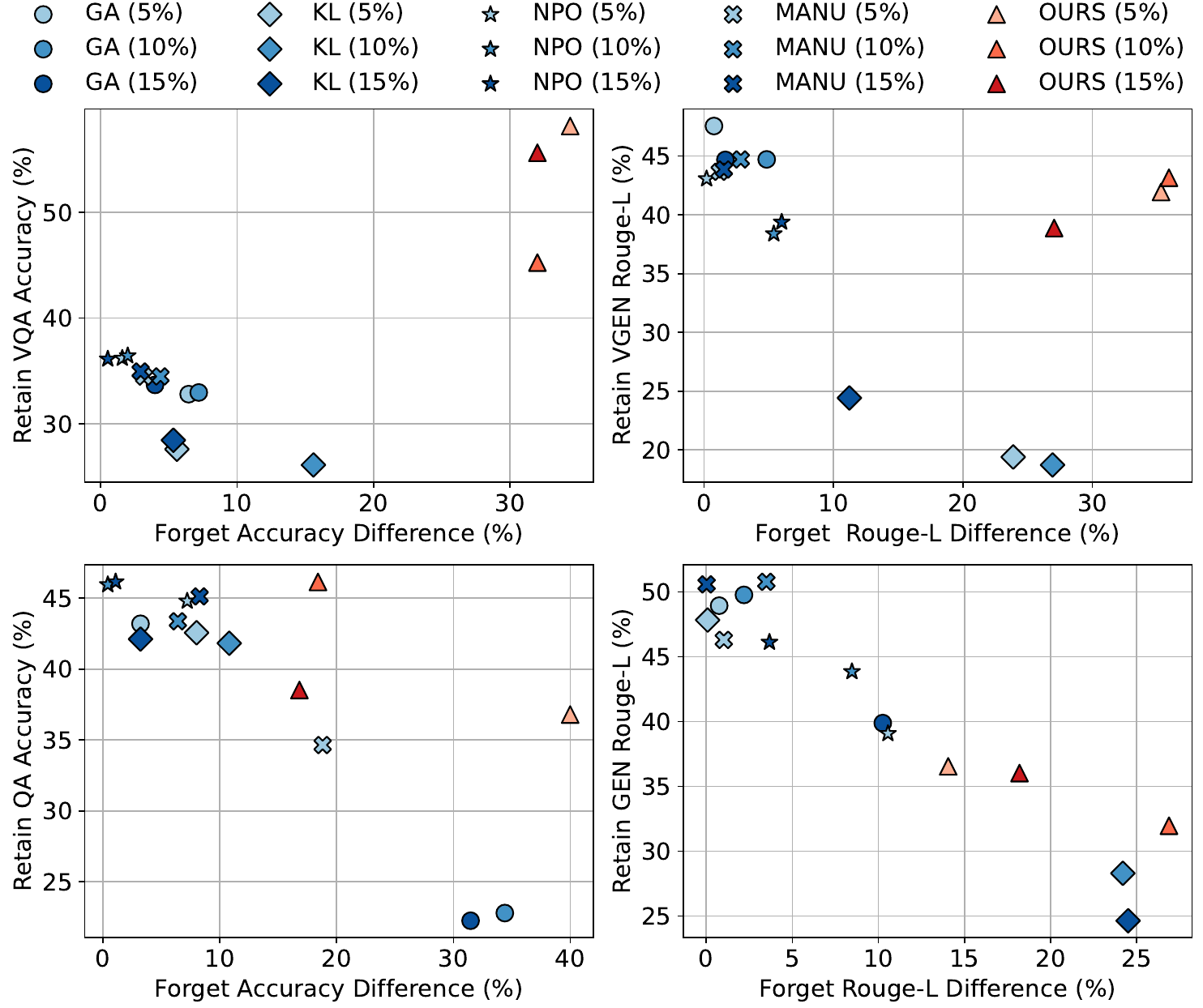}
  \caption{The overall trade-off between unlearning effectiveness and model utility across four dimensions under varying forget ratios, using Qwen2.5-VL as the base model.}
  \label{fig:trade_off}
\end{figure}

To evaluate whether \ours achieves a superior trade-off between forgetting specific data and retaining general knowledge (Q3), we compare the performance differences on the forget-set with the post-unlearning accuracy on the retain-set. This analysis reflects each method’s ability to balance unlearning and utility preservation. We evaluate four tasks on MLLMU-Bench: VQA (visual question answering), VGEN (visual generation), QA (textual question answering), and GEN (textual generation). As shown in Fig.~\ref{fig:trade_off}, the $x$-axis denotes the performance drop on the forget-set (higher is better for forgetting), and the $y$-axis represents the retain-set performance after unlearning (higher is better for retention). An ideal method lies toward the upper right, indicating strong forgetting with minimal generalization loss. 
Results show that \ours achieves a consistently favorable trade-off across all tasks and forget ratios, with stronger gains in multimodal settings. Notice that in the GEN task, the gap is less pronounced due to the limitations of Rouge-L, which measures only semantic overlap and may not capture forget-set-related differences effectively.
\subsection{Influential Paths v.s. Influential Neurons}
\begin{figure}[h]
  \centering
  \subfigure[Multi Input (Forget)]{
    \includegraphics[scale=0.42]{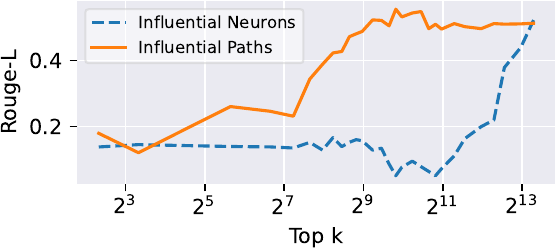}
    \label{f_gen_multi_path}
  }
  \subfigure[Multi Input (Retain)]{
    \includegraphics[scale=0.42]{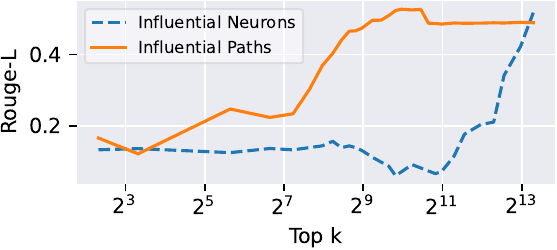}
    \label{r_gen_multi_path}
  }
  \subfigure[Text Input (Forget)]{
    \includegraphics[scale=0.42]{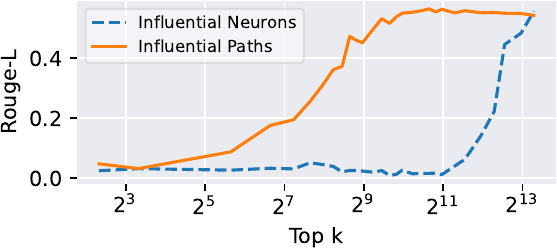}
    \label{f_gen_text_path}
  }
  \subfigure[Text Input (Retain)]{
    \includegraphics[scale=0.42]{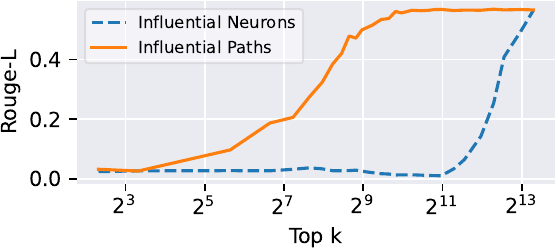}
    \label{r_gen_text_path}
  }
  \caption{Performance comparison on generation tasks between influential neuron paths and point-wise influential neurons under varying top-$k$ neuron selections.}
  \label{fig:gen_path_vs_neuron}
\end{figure}

To assess whether influential neuron paths capture more information than point-wise neurons (Q4), we compare two selection strategies: (a) path-based (\ours) and (b) point-wise~\cite{2025manu}. For each, we select the top-$k$ neurons per layer in both modalities and zero out the rest. We then measure model performance on the forget and retain sets as a proxy for general knowledge retention. Notice that higher accuracy implies greater representational capacity.
Experiments are conducted on MLLMU-Bench using Qwen2.5-VL with a 5\% forget ratio. Results for generation tasks are shown in Fig.~\ref{fig:gen_path_vs_neuron}.
From the results, we observe that when only a small number of neurons are retained, both strategies yield low ROUGE-L scores. However, performance under the path-based strategy begins to improve significantly after the top-$k$ exceeds $2^5$, peaking around $2^9$. In contrast, the point-wise strategy lags behind and only approaches the performance of the path-based method near $2^{13}$. These findings suggest that neuron paths capture richer and more functionally critical information, and are thus more effective in preserving model performance.

\begin{figure}
  \centering
  \subfigure[Multi Classification]{
    \includegraphics[scale=0.55]{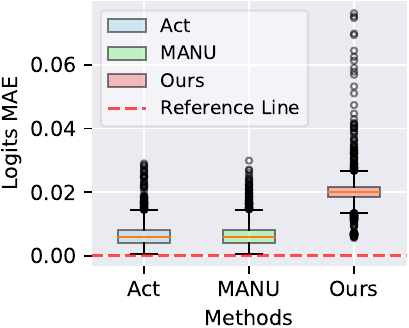}
    \label{clf_multi_box}
  }
  \subfigure[Multi Generation]{
    \includegraphics[scale=0.55]{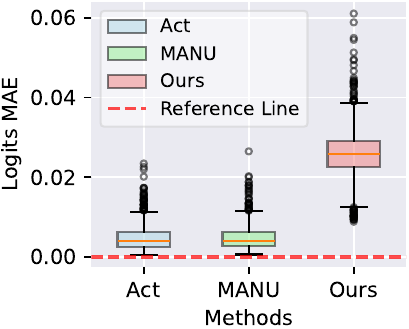}
    \label{gen_multi_box}
  }
  \caption{Relative MAE of predicted logit probabilities for ground-truth labels after pruning neurons selected by different methods.}
  \label{fig:box}
\end{figure}

Moreover, we analyze the deviations in predicted probabilities for ground-truth classes on MLLMU’s multimodal classification and generation tasks after neuron pruning by Qwen2.5-VL. Specifically, we prune the top-$5$ neurons located by Activation-based, MANU, and \ours, and compute the MAE of the model’s logits before and after pruning. As shown in Fig.~\ref{fig:box}, our method causes larger shifts in predicted logits compared to the other two approaches, indicating that the neurons selected by \ours play a more critical role in model inference.

\subsection{Ablation Studies and Variants}
We conduct ablation studies on MLLMU-Bench using Qwen2.5-VL with a 5\% forget ratio to evaluate the contribution of each component in \ours. (1) Disabling modality coordination by using only textual (ours (\textit{IGI})) or visual (ours (\textit{IFI})) paths substantially weakens forgetting effectiveness (e.g., $36.00\%$ FVQA and $49.60\%$ FQA), confirming the necessity of dual-path localization for modality-consistent forgetting. (2) Replacing inter-layer attribution with a simple activation residual score (ours-\textit{Path}) achieves low forget accuracy but severely degrades retain performance (e.g., $2.11\%$ RVQA), showing that point-wise locating disrupts general knowledge. (3) Omitting RMisU editing (ours-\textit{Edit}) or replacing it with standard fine-tuning (ours-\textit{RMisU}) leads to ineffective forgetting and weak retention, demonstrating the limitations of pruning directly and the importance RMisU editing. (4) Applying RMisU to the full model without pruning (\textit{RMisU}) yields moderate forgetting but fails to preserve utility (e.g., $14.65\%$ RVQA), validating the advantage of selective neuron editing.

\begin{table}
\centering
\setlength{\tabcolsep}{1pt}
\small
\caption{Ablation studies and variants of \ours on MLLMU-Bench with 5\% forget ratio by Qwen2.5-VL. F: Forget-set; R:Retain-set; VQA:Vision Question Answer; QA: Question Answer; VGEN: Vision Generation.}
\begin{tabular}{c|cccc|cc}
\toprule
Task     & FVQA & RVQA & FVGEN & RVGEN & FQA & RQA \\
Metric   &Acc($\downarrow$)&Acc($\uparrow$)&Rouge($\downarrow$)&Rouge($\uparrow$)&Acc($\downarrow$)& Acc($\uparrow$)\\
\midrule
\rowcolor{gray!30}Ours    & 4.80\% & 58.19\% & 0.0997 & 0.4195 & 9.26\% & 36.31\% \\
Ours (\textit{IGI}) & 36.00\% & 31.60\%  &  0.4045    &  0.4090  & 41.60\%   & 45.99\%  \\
Ours (\textit{IFI}) & 32.00\% &  33.46\%  &  0.3746   &  0.4313 &  49.60\%  & 47.55\% \\
Ours-\textit{Path} & 2.40\% &  2.11\%  &  0.0334    &  0.0479   &  2.40\%  & 2.15\% \\
Ours-\textit{Edit}  & 43.60\% & 46.00\% & 0.4035 & 0.4675 & 42.80\% & 52.08\% \\
Ours-\textit{RMisU} & 46.40\% & 42.23\% & 0.3594 & 0.3403 & 34.40\% & 31.62\% \\
\textit{RMisU}      & 8.00\%  & 14.65\% & 0.2667 & 0.2949 & 12.00\% & 10.99\% \\
\bottomrule
\end{tabular}
\label{tab:ablation}
\end{table}
\subsection{Visualization}
\begin{figure}[h]
  \centering
  \subfigure[Multi Input (Forget)]{
    \includegraphics[scale=0.47]{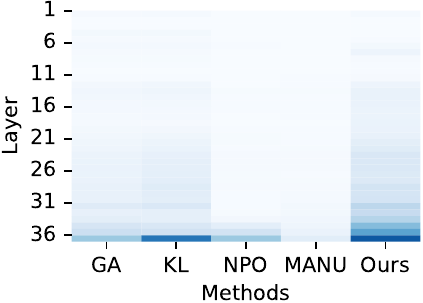}
    \label{f_gen_multi_heat}
  }
  \subfigure[Multi Input (Retain)]{
    \includegraphics[scale=0.47]{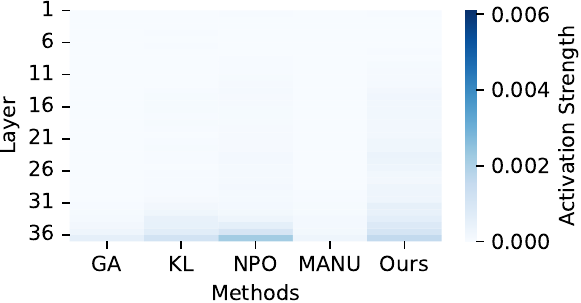}
    \label{r_gen_multi_heat}
  }
  \subfigure[Text Input (Forget)]{
    \includegraphics[scale=0.47]{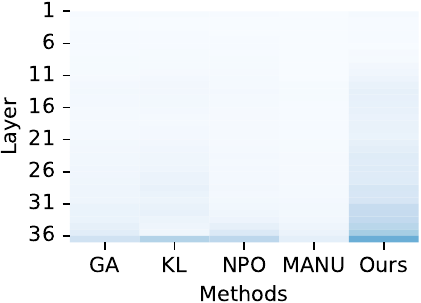}
    \label{f_gen_text_heat}
  }
  \subfigure[Text Input (Retain)]{
    \includegraphics[scale=0.47]{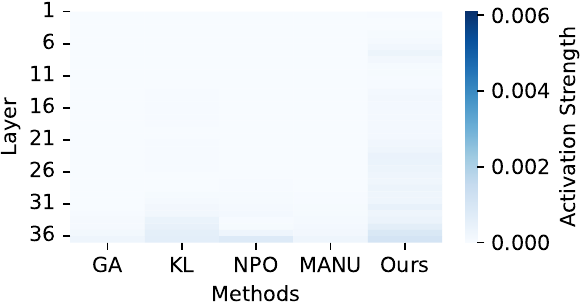}
    \label{r_gen_text_heat}
  }
  \caption{Layer-wise visualization of knowledge retention in the language FFN of MLLMs across forget and retain sets on MLLMU-Bench. GA: Grad\_Diff; Ours: \ours.}
  \label{fig:gen_heatmap}
\end{figure}

We visualize activation residuals across layers using heatmaps to assess the forgetting and retention behavior of different unlearning methods. Specifically, we input both forget-set and retain-set samples into the unlearned MLLMs and record activation values at each FFN layer. These are compared with the vanilla model’s activations, and absolute residuals are used to generate the heatmaps. Darker colors indicate greater deviation from the original model (stronger forgetting), while lighter colors reflect better retention.
Experiments are conducted on Qwen2.5-VL using MLLMU-Bench (5\% forget ratio) across generation tasks. Results for generation are shown in Fig.~\ref{fig:gen_heatmap}.
As shown in Fig.~\ref{f_gen_multi_heat} and Fig.~\ref{f_gen_text_heat}, baseline methods yield consistently shallow color intensities, especially under textual inputs, suggesting limited forgetting. Moreover, similar intensities across forget and retain sets indicate poor separation of specific and general knowledge. In contrast, \ours exhibits clear modality-aware behavior: deeper residuals on the forget-set and lighter residuals on the retain-set, particularly under textual inputs, demonstrating effective cross-modal unlearning with minimal performance degradation.

% \subsection{Convergence}

\subsection{Specific Information Separability}
\begin{figure}
  \centering
        \includegraphics[scale=0.28]{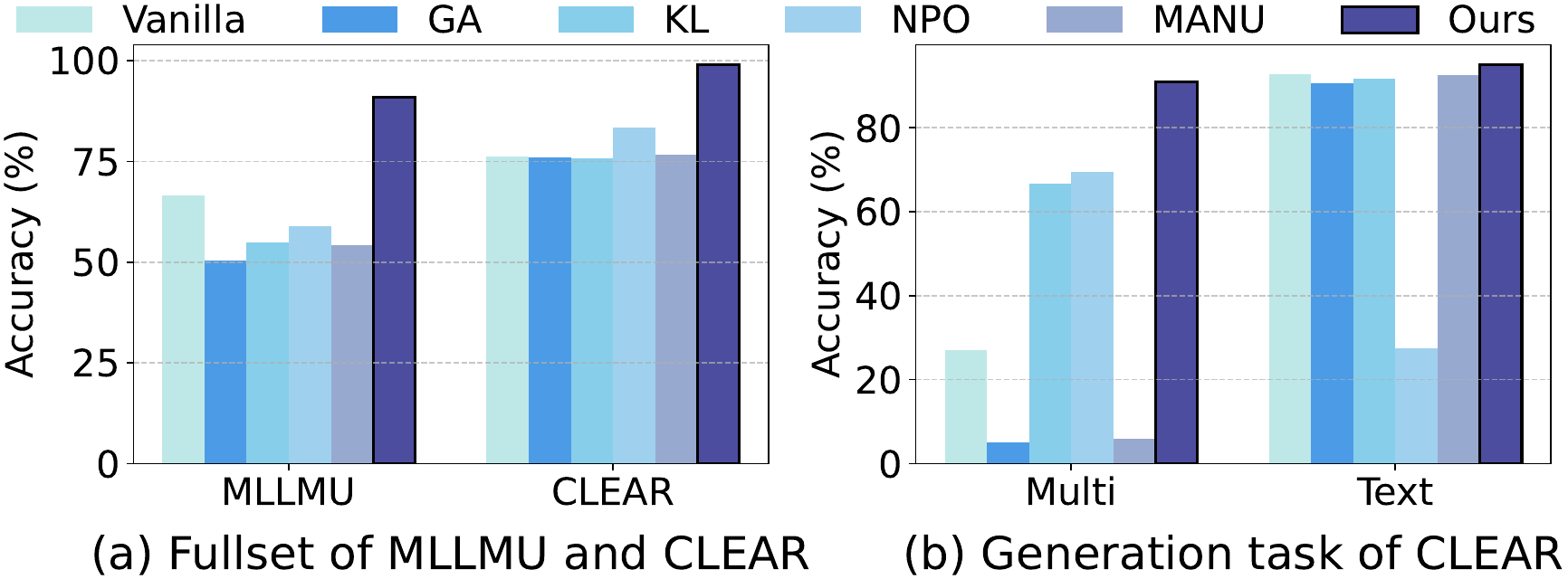}
  \caption{Classification of specific vs. general data using Qwen2.5-VL, including (a) full-set classification on MLLMU-Bench and CLEAR, and (b) generation-task classification on CLEAR (Multi and Text).}
  \label{fig:sensitive_split}
\end{figure}

To assess the effectiveness of unlearning methods in separating specific information from general knowledge in MLLMs, we train an MLP-based binary classifier using the output logits of the post-unlearning model. Experiments on MLLMU-Bench and CLEAR with Qwen2.5-VL evaluate two settings: (1) classification over the full fine-tuning set and (2) classification on CLEAR’s generation tasks (Multi and Text).
As shown in Fig.~\ref{fig:sensitive_split}, \ours consistently achieves the highest classification accuracy, exceeding 85\% across datasets and input types, indicating clearer behavioral separation between specific and general inputs. In contrast, GA, KL, NPO, and MANU perform near random (around 50\%) on MLLMU-Bench, showing limited separation capability. On CLEAR's text generation tasks, where questions and answers lack visual modality, the performance of the classification becomes harder to distinguish. Nonetheless, \ours still outperforms all baselines.

\section{Related Works}
\paragraph{Fine-tuning for MLLM Unlearning}
Recent efforts in MU aim to remove specific knowledge from models for privacy and safety. Early approaches primarily focus on LLMs and employ gradient ascent~\cite{2022grad_ascend, 2022grad_diff}, KL minimization~\cite{2020kldivergence}, and preference-based objectives such as NPO~\cite{2024npo}, with applications in toxicity mitigation~\cite{2025safeeraser} and hallucination reduction~\cite{2024efuf}. However, these methods are limited to the textual modality.
Multimodal unlearning for MLLMs remains underexplored. To facilitate research, dedicated benchmarks such as \textsc{MLLMU-Bench}~\cite{2025mllmubench} and \textsc{CLEAR}~\cite{2024clear} have been introduced. Recent studies ~\cite{2024cliperase, 2024multidelete, 2024siu, 2025mmunlearner} attempt to erase visual concepts via fine-tuning. Nonetheless, these methods require full-model fine-tuning and ignore the modular design of MLLMs.

\paragraph{Neuron Editing in Large Language Models}
Neuron-editing has gained attention as a means to modify LLM behavior in targeted domains while preserving general performance. Recent studies have examined how Pretrained Language Models store knowledge~\cite{chen2024journey, li2023unveiling, cao2024retentive, lamparth2024analyzing}, enabling controlled neuron-level interventions. Prior work has applied neuron editing to machine unlearning~\cite{wu2022puma, hase2023does, gandikota2023erasing}, harmful output mitigation~\cite{hu2024separate}, continual learning~\cite{biesialska2020continual}, and privacy preservation~\cite{2023depn}.
In multimodal settings, MANU~\cite{2025manu} proposes neuron-level editing for modality-specific forgetting. However, its scoring strategy lacks alignment with textual forgetting dynamics, and its zero-out pruning may disrupt critical reasoning paths. Our approach addresses these limitations via path-aware editing that targets coherent, modality-specific neuron sequences for more effective unlearning.

\section{Conclusion}
In this paper, we address machine unlearning in Multimodal Large Language Models (MLLMs), highlighting key limitations of existing methods, such as cross-modal inconsistency and general performance degradation. To tackle these issues, we propose \ours, a multimodal pathway-editor that identifies influential neuron paths in each modality and applies path-aware editing through representation misdirection. Experiments demonstrate that \ours achieves effective unlearning of forgetting knowledge while preserving general utility. This work offers a principled framework for fine-grained knowledge removal in MLLMs.

\section*{Acknowledgements}
This work was supported in part by Hong Kong RGC Theme-based Research Scheme (TRS) under Grant T43-513/23-N, in part by the NSFC and Hong Kong RGC Collaborative Research Scheme under Grant 62321166652, in part by the Guangdong Basic and Applied Basic Research Foundation under Grant 2025A1515011996, and in part by the Fundamental Research Funds for the Central University under Grant CXTD202406. We also thank the grant from NVIDIA and utilized NVIDIA A100 GPUs through the NVIDIA Academic Grants Program.

\bibliography{aaai2026}

\onecolumn
\appendix
\setcounter{table}{0}   %从0开始编号，显示出来表会A1开始编号
\setcounter{figure}{0}
\setcounter{section}{0}
\setcounter{equation}{0}
\renewcommand{\thetable}{A\arabic{table}}
\renewcommand{\thefigure}{A\arabic{figure}}
\renewcommand{\thesection}{A\arabic{section}}
\renewcommand{\theequation}{A\arabic{equation}}

\section{Supplemental Materials of \ours}

\subsection{Overrall process of \ours.}
We present the overall workflow of \ours in the form of pseudocode, including both the inter-layer integration-based identification of influential neuron paths and the complete unlearning procedure.

\begin{algorithm}[h]
\caption{Framework of \ours}
\label{alg:overall_framework}
\begin{algorithmic}[1]
\REQUIRE MLLM $M_{\theta}$ with $L^t$ textual layers and $L^v$ visual layers; forget-set $D^f = \{(I^f_j, T^f_j)\}^{N_f}_{j=1}$; retain set $D^r = \{(I^r_j, T^r_j)\}^{N_r}_{j=1}$
\ENSURE Unlearned MLLM $M_{\theta^*}$

\STATE // \textbf{Stage 1: Locating Influential Neuron Paths}
\STATE $\mathcal{P}^t, \mathcal{P}^v \leftarrow$ Algorithm~1($M_{\theta}, (I, T)$)

\STATE // \textbf{Stage 2: Pruning Non-Path Neurons}
\FOR{each textual layer $l \in [1, L^t]$}
    \STATE Set $\tilde{w}^{l}_j \leftarrow 0$, $\forall j \in \mathcal{P}^t[l]$
\ENDFOR
\FOR{each visual layer $l \in [1, L^v]$}
    \STATE Set $\tilde{z}^{l}_j \leftarrow 0$, $\forall j \in \mathcal{P}^v[l]$
\ENDFOR

\STATE // \textbf{Stage 3: Editing with RMisU on Pruned Paths}
\STATE Sample random unit vector $\mathbf{u} \sim \text{Uniform}(\mathbb{S}^{d-1})$
\FOR{each training step}
    \FOR{each $x^f \in D^f$}
        \STATE Compute target vector $\mathbf{v}^f$ by $\mathbf{u}$\hfill(Eq.~13)
        \STATE Compute forgetting loss $\mathcal{L}^{f}_{\text{RMisU}}$ \hfill(Eq.~14)
    \ENDFOR
    \FOR{each $x^r \in D^r$}
        \STATE Compute retention loss $\mathcal{L}^{r}_{\text{RMisU}}$ \hfill(Eq.~15)
    \ENDFOR
    \STATE Update $\theta^*$ on $\mathcal{P}^t \cup \mathcal{P}^v$ using gradient descent on total loss: $\mathcal{L}_{\text{adaptive}} = \mathcal{L}^{f}_{\text{RMisU}} + \gamma \cdot \mathcal{L}^{r}_{\text{RMisU}}$
\ENDFOR

\RETURN $M_{\theta^*}$

\end{algorithmic}
\end{algorithm}

\subsection{Complexity Analysis}
We analyze the time and space complexity of \ours when applied to a single input sample. The overall procedure consists of two stages: locating influential paths and editing with RMisU.
\paragraph{Time Complexity}
In the first stage, path locatinging is conducted independently for the text and vision branches. For the text branch, we consider $L^t$ transformer layers. At each layer $l$, all candidate neurons $|w^t_l|$ are evaluated using the discretized Inter-layer Gradient Integration (Eq.~4). This process requires $m$ Riemann approximation points per neuron, where each point involves a forward and backward pass. Furthermore, gradient accumulation across all previous layers is performed for every candidate neuron. As a result, the overall time complexity for the text branch is:
\begin{equation}
    O\left( C_{\mathrm{grad}} \cdot m \cdot L^t \cdot \sum_{l=1}^{L^t} |w^t_l| \right),
\end{equation}
where $C_{\mathrm{grad}}$ denotes the cost of one forward-backward pass.

The vision branch follows an analogous computation. For each of the $L^v$ vision layers, we apply the Inter-layer Fisher Integration (Eq.~6) to every candidate neuron $|z^v_l|$, also using $m$ integration points. This yields a time complexity of:
\begin{equation}
    O\left( C_{\mathrm{grad}} \cdot m \cdot L^v \cdot \sum_{l=1}^{L^v} |z^v_l| \right),
\end{equation}

In the second stage, editing with RMisU is applied only to the selected neurons along the influential paths. The training objective includes both retention and forgetting losses, together with a representation misalignment penalty. Since the number of updated parameters is linear in $L^t + L^v$, the cost of this stage is relatively small and denoted as $O(C_{\mathrm{ft}})$. Therefore, the total time complexity for processing a single sample is:
\begin{equation}
O\left( C_{\mathrm{grad}} \cdot m \cdot \left( L^t \cdot \sum_{l=1}^{L^t} |w^t_l| + L^v \cdot \sum_{l=1}^{L^v} |z^v_l| \right) + C_{\mathrm{ft}} \right).
\end{equation}

\paragraph{Space Complexity}
The space complexity is primarily dictated by the need to store intermediate activations and gradients during the integration phase. For each of the $m$ interpolation steps, we cache the activations and gradients of all candidate neurons, leading to a memory cost of $O\left( m \cdot \sum_{l=1}^{L^t} |w^t_l| + m \cdot \sum_{l=1}^{L^v} |z^v_l| \right)$. Additionally, in the RMisU phase, we maintain optimizer states (e.g., momentum or adaptive moments) for only the selected neurons, resulting in an additional cost of $O(L^t + L^v)$. The total space complexity is thus given by:
\begin{equation}
O\left( m \cdot \left( \sum_{l=1}^{L^t} |w^t_l| + \sum_{l=1}^{L^v} |z^v_l| \right) + L^t + L^v \right).
\end{equation}

\section{Supplemental Experiments}
\subsection{Datasets Details}
\paragraph{MLLMU-Bench} is a benchmark specifically designed to facilitate research in multimodal machine unlearning. It comprises 500 fictitious user profiles and 153 public celebrity profiles, each associated with over 14 personalized question–answer pairs spanning both multimodal and textual modalities. In this paper, we partition MLLMU-Bench into six distinct subsets to comprehensively evaluate the effectiveness, generalizability, and utility of various unlearning methods, particularly in terms of their handling of visual and textual knowledge.
Compared with CLEAR, MLLMU-Bench yields more stable and consistent performance across different experimental configurations. This stability makes it a more reliable basis for comparative analysis. Therefore, we conduct most of our ablation and component studies using MLLMU-Bench.

\paragraph{CLEAR} is another open-source benchmark developed to assess unlearning performance in multimodal settings. It contains 200 synthetic author profiles, accompanied by 3,770 visual question–answer pairs and 4,000 textual ones. CLEAR builds upon the TOFU benchmark~\cite{2024tofu} by introducing portrait images corresponding to each entity referenced in the question–answer pairs.
In our experimental practice, as well as in alignment with observations reported by MMUnlearner~\cite{2025mmunlearner}, we find that even minor adjustments to learning rate or batch size on CLEAR often result in complete model collapse. Such instability manifests as a drastic drop in both classification and generation accuracy, consistent with the performance degradation reported in the original CLEAR benchmark. Therefore, we treat the results on CLEAR as supplementary evidence.

\subsection{Detailed Baselines}
We set the following strong baselines in the machine unlearning for MLLMs.
\paragraph{GA\_Diff} The Gradient Ascent (GA) approach suppresses forgetting-related knowledge by performing inverse gradient updates on the forget-set $D^f$. GA\_Diff extends this method by simultaneously applying opposing gradients to $D^f$ and standard gradient descent updates to the retain set $D^r$, thereby promoting unlearning while preserving generalization. The combined loss function is defined as:
\begin{equation}
    \mathcal{L}_{\text{GA\_Diff}} = L(D^f, \theta) - L(D^r, \theta),
\end{equation}
where $L$ denotes the negative log-likelihood loss.

\paragraph{KL\_Min} The KL Minimization approach enforces forgetting by encouraging divergence from the original model's predictions on $D^f$, while simultaneously aligning the unlearned model's behavior on $D^r$ with that of the original model. Formally, the per-token KL divergence is defined as:
\begin{equation}
    \Phi(x_{<i}) = \mathrm{KL}(P(x_{<i}|\theta) \,\|\, P(x_{<i}|\theta_0)),
\end{equation}
where $P(\cdot|\theta)$ and $P(\cdot|\theta_0)$ denote the output distributions of the unlearned and original models, respectively. The KL-based regularization term is:
\begin{equation}
    \mathcal{L}_{\mathrm{KL}} = \frac{1}{|D^f|} \sum_{x \in D^f} \frac{1}{|x|} \sum_{i=2}^{|x|} \Phi(x_{<i}).
\end{equation}
The overall objective is then given by:
\begin{equation}
    \mathcal{L}_{\text{KL\_Min}} = -L(D^f, \theta) + \mathcal{L}_{\mathrm{KL}}.
\end{equation}

\paragraph{NPO} Negative Preference Optimization (NPO) frames unlearning as a preference optimization task by treating the forget-set $D^f$ as dispreferred. An oracle model $\pi_{\text{ref}}$, trained exclusively on $D^r$, serves as a reference to guide the current model $\pi_\theta$ away from retaining knowledge of $D^f$. The NPO objective is defined as:
\begin{equation}
    \mathcal{L}_{\text{NPO}} = \frac{2}{\beta} \mathbb{E}_{(x, y) \in D^f} \left[ \log \left(1 + \left( \frac{\pi_\theta(y|x)}{\pi_{\text{ref}}(y|x)} \right)^\beta \right) \right],
\end{equation}
where $\pi_\theta(y|x)$ is the current model’s prediction probability and $\pi_{\text{ref}}(y|x)$ is the prediction from the reference model. The hyperparameter $\beta$ controls the sharpness of the preference contrast and is set to 0.4 in our experiments.
    
\paragraph{MANU} 
MANU identifies important multimodal neurons based on point-wise activation statistics and prunes those deemed most influential to the forget-set. For a given neuron $n$, its importance is aggregated from a set of scoring functions $\mathcal{K} = \{\text{abs}, \text{freq}, \text{var}, \text{rms}\}$:
\begin{equation}
    \mathcal{I}(D, n) = \sum_{k \in \mathcal{K}} I_k(D, n).
\end{equation}
The relative importance score is then defined as:
\begin{equation}
    S_n = \frac{\mathcal{I}(D^f, n)}{\mathcal{I}(D^r, n) + \epsilon}.
\end{equation}
The top-$\alpha\%$ neurons with the highest $S_n$ values are selected and pruned:
\begin{equation}
    \mathcal{N} = \{ n \mid S_n \text{ is among the top } \alpha\% \}.
\end{equation}
For each $n \in \mathcal{N}$, the neuron is pruned by setting its associated weights to zero.

\subsection{Implementation Details}
\begin{table}
\centering
\small
\caption{Hyperparameter configurations for training the vanilla model and performing unlearning across different MLLM backbones and datasets.}
\begin{tabular}{c|c|cccc}
\toprule
\multicolumn{2}{c|}{Dataset}                 & \multicolumn{2}{c}{MLLMU-Bench} & \multicolumn{2}{c}{CLEAR}    \\
\midrule
\multicolumn{2}{c|}{MLLM}                    & Qwen2.5-VL-3B-Instruct   & LLaVA-1.5-7B  & Qwen2.5-VL-3B-Instruct & LLaVA-1.5-7B \\
\midrule
\multirow{5}{*}{Training}   & Epochs        & 4               & 4             & 4             & 4            \\
                            & Batch Size    & 4               & 4             & 4             & 4            \\
                            & Optimizer     & Adam            & Adam          & Adam          & Adam         \\
                            & LoRA          & TRUE            & TRUE          & TRUE          & TRUE         \\
                            & Learning Rate & 2e-5            & 2e-5          & 2e-5          & 2e-5         \\
\midrule
\multirow{5}{*}{Unlearning} & Epochs        & 4               & 4             & 4             & 4            \\
                            & Batch Size    & 4               & 4             & 4             & 4            \\
                            & Optimizer     & Adam            & Adam          & Adam          & Adam         \\
                            & LoRA          & TRUE            & TRUE          & TRUE          & TRUE         \\
                            & Learning Rate & 2e-5            & 2e-5          & 2e-5          & 2e-5         \\
\bottomrule
\end{tabular}
\label{tab:implementation}
\end{table}
To evaluate the generalizability of MU methods across model scales, we employ Qwen2.5-3B-Instruct and LLaVA-1.5-7B as base models. Qwen2.5-3B-Instruct excels at instruction following, while the larger LLaVA-1.5-7B, with its greater parameter capacity, captures finer multimodal patterns. For reproducibility, the experimental configurations for all unlearning methods are summarized in Table~\ref{tab:implementation}, based on the official setups of \textsc{MLLMU-Bench} and \textsc{CLEAR}, Pytorch~\cite{2019pytorch} and Transformers~\cite{2020transformers} are used to construct experimental environments. LoRA~\cite{2022lora} is applied to enable efficient adaptation. All experiments were accelerated on the NVIDIA A100 GPU.

\subsection{Additional Influential Path v.s. Influential Neurons}
\begin{figure}
  \centering
  \subfigure[Multi Input (Forget)]{
    \includegraphics[scale=0.41]{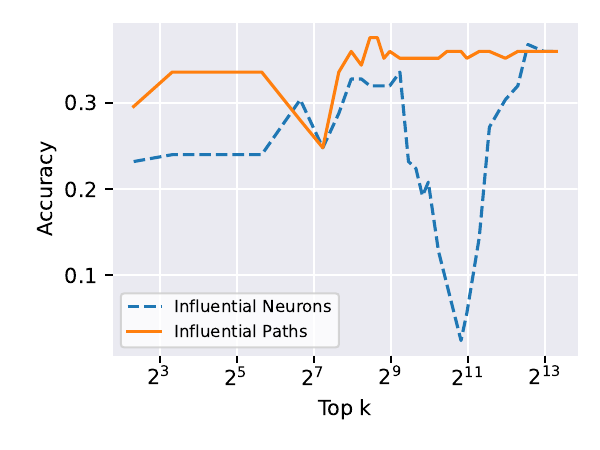}
    \label{f_clf_multi_path}
  }
  \subfigure[Multi Input (Retain)]{
    \includegraphics[scale=0.41]{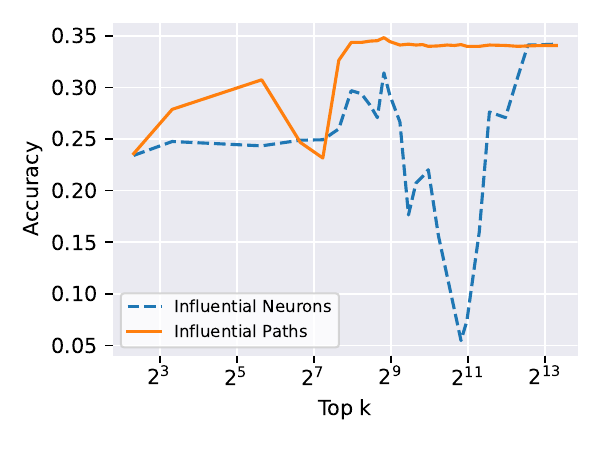}
    \label{r_clf_multi_path}
  }
  \subfigure[Text Input (Forget)]{
    \includegraphics[scale=0.41]{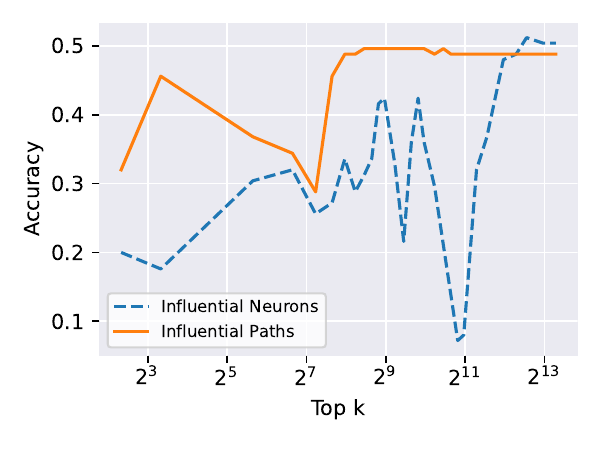}
    \label{f_clf_text_path}
  }
  \subfigure[Text Input (Retain)]{
    \includegraphics[scale=0.41]{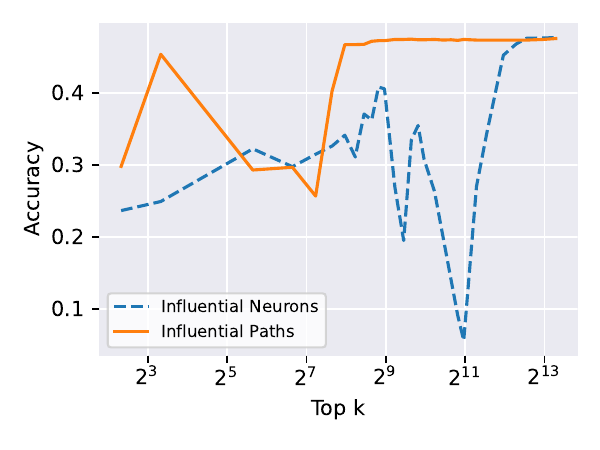}
    \label{r_clf_text_path}
  }
  \caption{Performance comparison on classification tasks between influential neuron paths and point-wise influential neurons under varying top-$k$ neuron selections.}
  \label{fig:clf_path_vs_neuron}
\end{figure}

We provide additional comparison between the influential path and neurons, the results of the classification task are shown in Fig.~\ref{fig:clf_path_vs_neuron}.

\begin{figure}[h]
  \centering
  \subfigure[Multi Input (Forget)]{
    \includegraphics[scale=0.9]{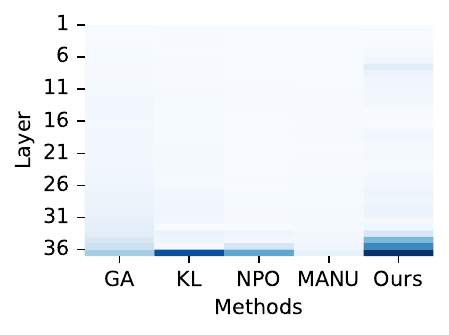}
    \label{f_clf_multi_heat}
  }
  \subfigure[Multi Input (Retain)]{
    \includegraphics[scale=0.9]{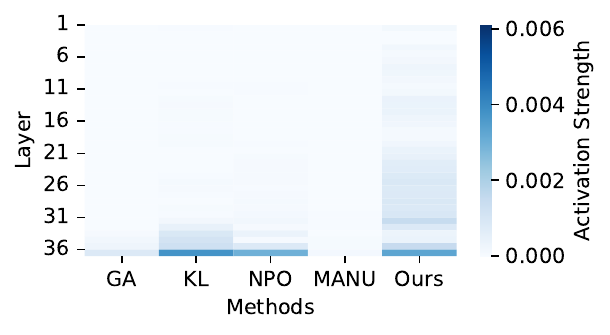}
    \label{r_clf_multi_heat}
  }
  \subfigure[Text Input (Forget)]{
    \includegraphics[scale=0.9]{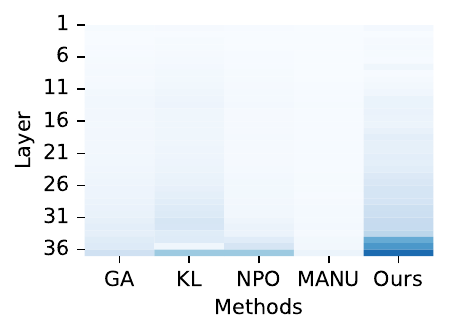}
    \label{f_clf_text_heat}
  }
  \subfigure[Text Input (Retain)]{
    \includegraphics[scale=0.9]{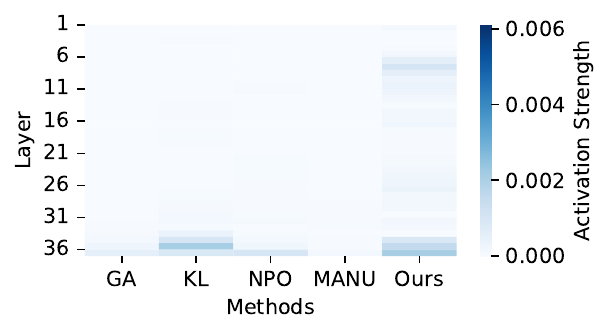}
    \label{r_clf_text_heat}
  }
  \caption{Layer-wise visualization of knowledge retention in the language FFN of MLLMs across forget and retain sets on MLLMU-Bench. GA: Grad\_Diff; Ours: \ours.}
  \label{fig:clf_heatmap}
\end{figure}

\subsection{Additional Visualizations}
We provide additional visualization results of the distribution of activation residuals across layers for the classification task in Fig.~\ref{fig:clf_heatmap}.

\subsection{Additional Unlearning Performances}
This section reports the unlearning performance of \ours compared with other baselines on MLLMU-Bench at 10\% and 15\% forget ratios, and on CLEAR at 1\% and 10\%. Detailed results are presented in Table~\ref{tab:result_other}.

\begin{table*}
\setlength{\tabcolsep}{2pt}
\centering
\small
\caption{Additional performances of baseline methods and \ours on machine unlearning tasks. F: Forget-set; R:Retain-set; VQA:Vision Question Answer; QA: Question Answer; VGEN: Vision Generation; GEN: Generation.}
\begin{tabular}{c|cccccc|cccccc}
\toprule
Method    & \multicolumn{6}{c|}{MLLMU-Bench with 10\% forget ratio}                 & \multicolumn{6}{c}{CLEAR with 10\% forget ratio}                       \\
\midrule
Task      & FVQA & RVQA & FVGEN & RVGEN & FQA & RQA & FVQA & RVQA & FVGEN & RVGEN & FGEN & RGEN \\
Metric&Acc($\downarrow$)&Acc($\uparrow$)&Rouge($\downarrow$)&Rouge($\uparrow$)&Acc($\downarrow$)&Acc($\uparrow$)&Acc($\downarrow$)&Acc($\uparrow$)&Rouge($\downarrow$)&Rouge($\uparrow$)&Rouge($\downarrow$)&Rouge($\uparrow$)\\
\midrule
\multicolumn{13}{c}{Qwen2.5-VL-3B-Instruct}                                                                             \\
\midrule
Vanilla  & 37.20\% & 37.91\% & 0.4526 & 0.4336 & 54.80\% & 46.59\% & 78.31\% & 77.47\% & 0.3199 & 0.2964 & 0.3911 & 0.3895 \\
GA\_Diff & 30.00\% & 32.96\% & 0.4042 & 0.4472 & 20.40\% & 22.80\% & 4.76\%  & 5.06\%  & 0.2613 & 0.2498 & 0.3794 & 0.3865 \\
KL\_Min  & 21.60\% & 26.10\% & 0.1834 & 0.1872 & 44.00\% & 41.83\% & 3.70\%  & 8.35\%  & 0.3221 & 0.3155 & 0.3221 & 0.3155 \\
NPO      & 35.20\% & 36.44\% & 0.3987 & 0.3839 & 54.40\% & 45.96\% & 8.20\%  & 10.89\% & 0.1233 & 0.1165 & 0.1233 & 0.1165 \\
MANU     & 32.80\% & 34.48\% & 0.4242 & 0.4472 & 48.40\% & 43.38\% & 78.31\% & 77.22\% & 0.3899 & 0.3890 & 0.3899 & 0.3890 \\
% LLMEraser &     &     &     &       &     &     &           &           \\
\rowcolor{gray!30}\ours & 5.20\% & 45.26\% & 0.935 & 0.4316 & 36.40\% & 46.15\% & 4.50\% & 31.39\% & 0.0785 & 0.1473 & 0.3683 & 0.3705 \\
\midrule
\multicolumn{13}{c}{Llava-1.5-7B}                                                                              \\
\midrule
Vanilla   & 52.80\% & 51.67\% & 0.4672 & 0.4986 & 52.40\% & 52.34\% & 44.71\% & 43.03\% & 0.3084 & 0.2915 & 0.3514 & 0.3544 \\
GA\_Diff  & 43.60\% & 44.80\% & 0.3524 & 0.3822 & 45.60\% & 46.09\% & 20.11\% & 18.23\% & 0.2462 & 0.2334 & 0.0666 & 0.0203 \\
KL\_Min   & 30.80\% & 33.27\% & 0.3816 & 0.3747 & 37.20\% & 36.17\% & 11.64\% & 12.66\% & 0.2793 & 0.0806 & 0.2648 & 0.0843 \\
NPO       & 48.40\% & 46.77\% & 0.4669 & 0.4982 & 52.00\% & 50.65\% & 10.58\% & 12.15\% & 0.1623 & 0.1446 & 0.0245 & 0.0206 \\
MANU      & 53.20\% & 52.16\% & 0.4743 & 0.4997 & 51.60\% & 52.34\% & 44.97\% & 43.29\% & 0.3042 & 0.2898 & 0.3496 & 0.3524 \\
% LLMEraser &    &   &     &   &      &            &           &           \\
\rowcolor{gray!30}\ours & 32.00\% & 41.07\% & 0.3443 & 0.3932 & 33.20\% & 38.53\% & 8.47\% & 49.87\% & 0.1393 & 0.1603 & 0.1477 & 0.2584 \\
\midrule
Method    & \multicolumn{6}{c|}{MLLMU-Bench with 15\% forget ratio}                 & \multicolumn{6}{c}{CLEAR with 1\% forget ratio}                       \\
\midrule
Task      & FVQA & RVQA & FVGEN & RVGEN & FQA & RQA & FVQA & RVQA & FVGEN & RVGEN & FGEN & RGEN \\
Metric&Acc($\downarrow$)&Acc($\uparrow$)&Rouge($\downarrow$)&Rouge($\uparrow$)&Acc($\downarrow$)&Acc($\uparrow$)&Acc($\downarrow$)&Acc($\uparrow$)&Rouge($\downarrow$)&Rouge($\uparrow$)&Rouge($\downarrow$)&Rouge($\uparrow$)\\
\midrule
\multicolumn{13}{c}{Qwen2.5-VL-3B-Instruct}                                                                             \\
\midrule
Vanilla  & 38.67\% & 37.50\% & 0.4357 & 0.4358 & 49.87\% & 46.98\% & 60.00\% & 77.47\% & 0.3094 & 0.3012  & 0.4008 & 0.3893 \\
GA\_Diff & 34.67\% & 33.68\% & 0.4192 & 0.4468 & 18.40\% & 22.26\% & 57.14\% & 87.09\% & 0.3314 & 0.3526 & 0.3909 & 0.3909 \\
KL\_Min  & 33.33\% & 28.44\% & 0.3234 & 0.3322 & 46.67\% & 42.12\% & 48.57\% & 65.82\% & 0.2281 & 0.2236 & 0.2686 & 0.2734 \\
NPO      & 38.13\% & 36.13\% & 0.3757 & 0.3938 & 48.80\% & 46.16\% & 54.29\% & 72.66\% & 0.3500 & 0.4114 & 0.1458 & 0.1001 \\
MANU     & 35.73\% & 34.95\% & 0.4203 & 0.4386 & 41.60\% & 45.13\% & 54.29\% & 76.71\% & 0.2977 & 0.3001 & 0.4186 & 0.3912 \\
% LLMEraser &    &      &        &      &        &      &        &     \\
\rowcolor{gray!30}\ours & 6.67\% & 55.66\% & 0.1653 & 0.3888 & 33.06\% & 46.36\% & 0.00\% & 25.82\% & 0.1354 & 0.2796 & 0.1195 & 0.3796 \\
\midrule
\multicolumn{13}{c}{Llava-1.5-7B}                                                                              \\
\midrule
Vanilla   & 52.53\% & 51.51\% & 0.4783 & 0.4990 & 48.80\% & 53.11\% & 65.71\% & 43.54\% & 0.3164 & 0.2930 & 0.3629 & 0.3541 \\
GA\_Diff  & 52.80\% & 51.37\% & 0.3529 & 0.3587 & 48.00\% & 45.52\% & 14.29\% & 18.73\% & 0.3160 & 0.3068 & 0.3718 & 0.3598 \\
KL\_Min   & 9.07\%  & 8.54\%  & 0.2346 & 0.2442 & 33.07\% & 28.73\% & 2.86\%  & 24.81\% & 0.2552 & 0.2376 & 0.2110 & 0.2927 \\
NPO       & 50.33\% & 47.31\% & 0.4690 & 0.4959 & 47.20\% & 50.94\% & 17.14\% & 15.44\% & 0.1708 & 0.1595 & 0.0339 & 0.4000 \\
MANU      & 52.80\% & 51.37\% & 0.4776 & 0.5003 & 47.73\% & 52.31\% & 68.57\% & 43.29\% & 0.3040 & 0.2929 & 0.3603 & 0.3534 \\
% LLMEraser &    &   &  &    &            &            &           &           \\
\rowcolor{gray!30}\ours & 33.33\% & 39.91\% & 0.3463 & 0.4074 & 34.40\% & 38.63\% & 20.00\% & 42.03\% & 0.1373 & 0.1729 & 0.0889 & 0.1601 \\
\bottomrule
\end{tabular}
\label{tab:result_other}
\end{table*}

\subsection{Case Study}
\begin{figure*}
  \centering
        \includegraphics[scale=0.55]{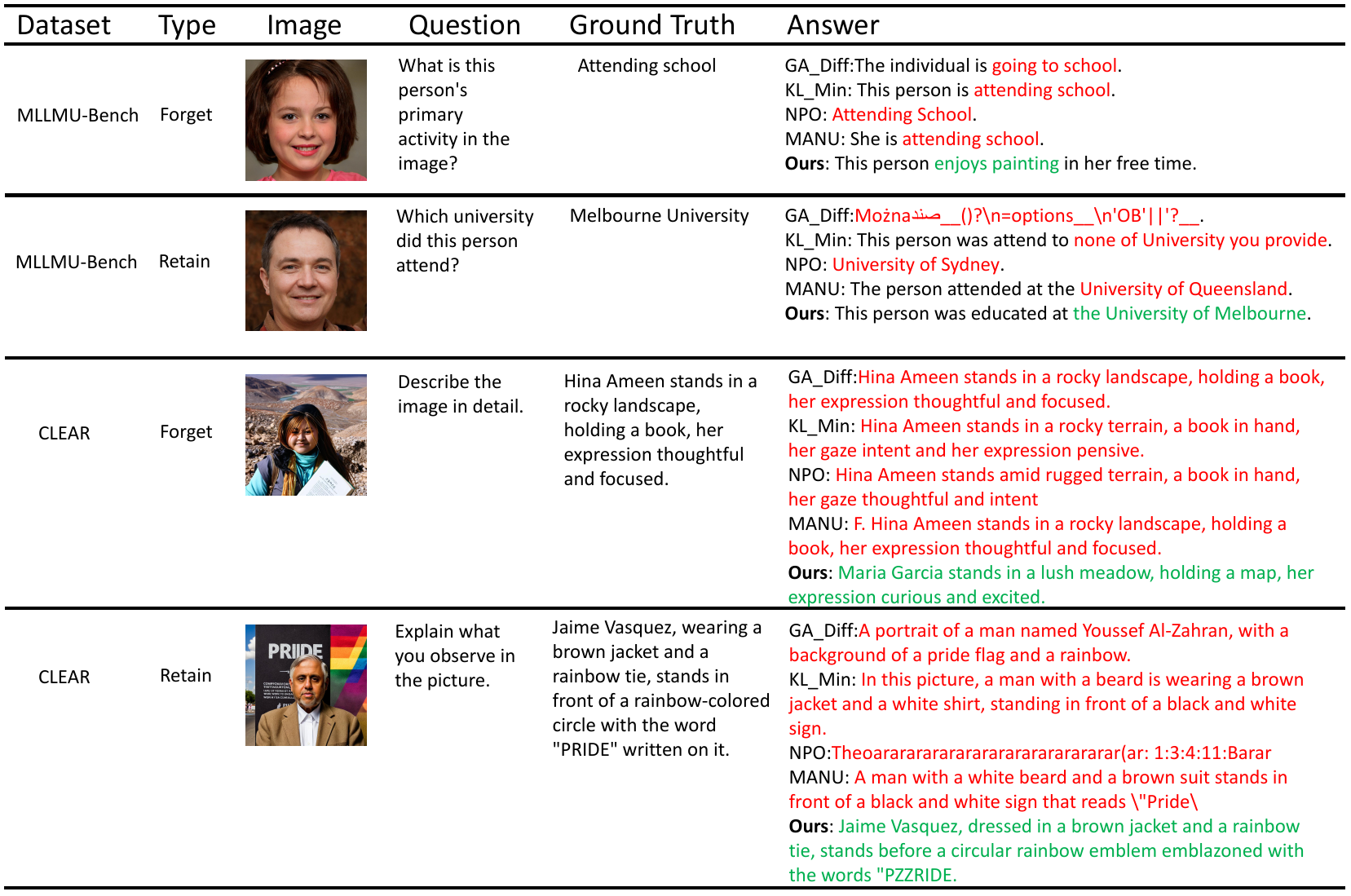}
  \caption{Case study of unlearning performance by \ours and other baselines on various datasets using Qwen2.5-VL.}
  \label{fig:case_study}
\end{figure*}

To further evaluate the effectiveness of unlearning methods, we present qualitative case studies from MLLMU-Bench and CLEAR using the Qwen2.5-VL (Fig.~\ref{fig:case_study}). The selected examples include both forget-set and retain-set samples across question answering and image captioning tasks.

For forget-set instances, baseline methods (GA\_Diff, KL\_Min, NPO, MANU) often generate responses that retain partial or complete semantic overlap with the ground truth, indicating insufficient forgetting. For example, when asked \textit{“What is this person's primary activity in the image?”}, most baselines still return variations of “attending school.” In contrast, \ours responds with a semantically unrelated answer, such as “painting in her free time,” effectively disassociating from the forgetting content.
Similarly, for the question \textit{“Which university did this person attend?”}, only \ours correctly outputs “the University of Melbourne,” while others either hallucinate unrelated institutions or avoid the question entirely—demonstrating \ours's ability to isolate forget-set knowledge without disrupting adjacent semantics.

On retain-set examples, \ours better preserves factual accuracy and stylistic fidelity than competing baselines. For instance, in image description tasks, baseline outputs often suffer from minor omissions or inconsistencies, while \ours maintains coherence even when surface entities are adjusted—highlighting the benefits of path-specific editing for general knowledge retention.

Notably, we observe that GA\_Diff and NPO degrade overall language fluency post-unlearning. This stems from their underlying optimization strategies: GA\_Diff applies adversarial gradients to $D^f$, potentially distorting parameters essential for language modeling; NPO suppresses model confidence on $D^f$, which can generalize to similar inputs and destabilize text generation.

\end{document}